\documentclass{article}
\usepackage{spconf,amsmath,graphicx,color,bm,hyperref}
\usepackage{booktabs}
\usepackage{multirow}
\usepackage{placeins}
\usepackage{cite}

\pdfoutput=1

\newcommand\textss[1]{\textsuperscript#1}

\newcommand{\namedref}[2]{\hyperref[#2]{#1~\ref*{#2}}}

\newcommand{\sectionref}[1]{\namedref{Section}{sec:#1}}
\newcommand{\tableref}[1]{\namedref{Table}{tab:#1}}
\newcommand{\figureref}[1]{\namedref{Figure}{fig:#1}}

\title{Hierarchical  Multitask Learning for \\ CTC-based speech recognition}
\name{Kalpesh Krishna\textss{1}, Shubham Toshniwal\textss{2}, Karen Livescu\textss{2}}
\address{\textss{1}University of Massachusetts, Amherst, USA
  \;\; \textss{2}TTI-Chicago, USA\\
 {\small \tt {kalpesh@cs.umass.edu~ \{shtoshni, klivescu\}@ttic.edu}
  }
  }

\begin{document}
\ninept
\maketitle
\begin{abstract}
  Previous work has shown that neural encoder-decoder speech recognition can be improved with hierarchical multitask learning, where auxiliary tasks are added at intermediate layers of a deep encoder.  
  We explore the effect of hierarchical multitask learning in the context of connectionist temporal classification (CTC)-based speech recognition, and investigate several aspects of this approach. 
  Consistent with previous work, we observe performance improvements on telephone conversational speech  recognition (specifically the Eval2000 test sets) when training a subword-level CTC model with an auxiliary phone loss at an intermediate layer.  
  We analyze the effects of a number of experimental variables (like interpolation constant and position of the auxiliary loss function),  performance in lower-resource settings, and the relationship between 
  pretraining and multitask learning. We observe that the hierarchical multitask approach improves over standard multitask training in our higher-data experiments, while in the low-resource settings standard multitask training works well.  The best results are obtained by combining hierarchical multitask learning and pretraining, which improves word error rates by 3.4\% absolute on the Eval2000 test sets.
\end{abstract}


\section{Introduction}
Modern automatic speech recognition (ASR) systems are increasingly moving toward neural ``end-to-end'' architectures that map audio directly to text, with no interpretable intermediate labels.  These models typically pass the input audio features through a multi-layer neural encoder, most often a recurrent neural network (RNN), to obtain a higher-level representation.  The output text is then obtained by passing this representation through an RNN-based decoder~\cite{cho2014learning} or using connectionist temporal classification (CTC)~\cite{graves2006connectionist}. This is in contrast to traditional ASR systems, which typically involve several separately trained components such as frame classifiers, acoustic models, lexicons, and language models.  These individual modules correspond to different levels of representation, such as triphone states, phonemes, graphemes or words.

There is strong evidence that intermediate encoder layers of end-to-end architectures, both for ASR and for other tasks, \textit{implicitly} learn intermediate representations between the input and the final output. End-to-end ASR systems appear to learn intermediate phonetic representations~\cite{belinkov2017analyzing}; language models learn syntactic representations in lower layers~\cite{PetersELMo2018}; and deep CNNs trained for image classification learn about curves, edges, and object parts in the image~\cite{zeiler2014visualizing}. 

Some recent work has introduced additional auxiliary loss functions at intermediate model layers and have found improvements on the primary task.  This is a form of multitask learning~\cite{caruana1997multitask}, but unlike most multitask learning, the secondary loss function is not applied at the final output layer.  For example, part-of-speech tag supervision at lower layers can improve performance of a neural syntactic chunker or tagger~\cite{sogaard2016deep}.  More recently, phonetic recognition and frame-level state classification losses applied to intermediate layer representations have been found to improve a character-level encoder-decoder recognizer on conversational telephone speech~\cite{toshniwal2017mtl}, and various levels of intermediate-layer subword losses improve a subword-based CTC recognizer~\cite{sanabria2018hierarchical}.

In this paper we investigate the benefit of a lower-layer phone-level CTC loss on a subword-level CTC model for speech recognition, trained on the Switchboard 300-hour training set~\cite{godfrey1992switchboard} and tested on the Eval2000 test sets~\cite{ld20022000}.  We investigate the effect of the mixing weight between the two CTC losses (\sectionref{interpolate}), as well as the choice of layer for the phone CTC loss (\sectionref{position}).  We also evaluate our models in lower-resource settings, using only a fraction of the training data (\sectionref{partial}).  We compare the hierarchical multitask approach to pretraining with the phone CTC loss and also combine the two approaches (\sectionref{pretrain}). Finally, we qualitatively compare the CTC alignments produced by the baseline and proposed models (\sectionref{alignment}). Throughout the paper, we also evaluate our model's performance on the auxiliary phonetic recognition task.
\section{Related Work}
Multitask learning (MTL) has been a common tool in machine learning for some time~\cite{caruana1997multitask}, and recent work has found that end-to-end neural speech recognition models can benefit from this approach~\cite{kim2017joint, rao2017hierarchical,toshniwal2017mtl}. 
Among the first work, to our knowledge, to use {\it hierarchical} multitask learning in speech recognition is that of Fern\'andez {\em et al.}~\cite{fernandez2007sequence}, which used a hierarchical CTC model, similar to ours although only two layers deep, with phoneme label prediction as an auxiliary task in training a spoken digit sequence recognizer.  
 Their experiments, with a fixed interpolation constant and phoneme prediction loss applied at the first layer of a two-layer deep RNN encoder, showed no improvement over the baseline.
Rao {\it et al.}~\cite{rao2017hierarchical} use a hierarchical CTC model for multi-accent speech recognition which is trained with the ASR loss applied at the topmost layer of a deep RNN encoder and a phoneme loss for different accents applied at a fixed intermediate layer with a fixed weight~\cite{rao2017hierarchical}.
Audhkhasi {\em et al.}~\cite{audhkhasi2017direct} experiment with different strategies to pretrain a CTC model with a phonetic loss and then continue training it in a hierarchical MTL framework combining the phonetic loss with the primary ASR loss. Their experiments found that pretraining alone worked as well as the combination of pretraining and multitask learning.  In our exploration of a larger space of pretraining + hierarchical multitask learning models, we do find a performance advantage with this combined approach (see \sectionref{pretrain}).
Toshniwal {\em et al.}~\cite{toshniwal2017mtl} experiment with hierarchical MTL in recurrent encoder-decoder models for ASR, and found that applying an auxiliary loss (a phonetic recognition loss, a frame-level state classification loss, or both) at an intermediate encoder layer improves performance over both a baseline model and standard MTL with all losses at the topmost layer. Sanabria {\em et al.}~\cite{sanabria2018hierarchical} used a hierarchical CTC model and varied the CTC vocabulary in their auxiliary tasks (character level, and a variety of subword units corresponding to different vocabulary sizes%\stcomment{added a bit more info}
), obtaining excellent final results. %300 subwords units, 1000 subwords units and so on). 

Our work here can be seen as an extension of these several past attempts to use variations of hierarchical multitask learning in neural speech recognition.  We contribute a  thorough investigation of
a number of previously unexplored aspects of the hierarchical MTL space, and find consistently larger improvements.
\section{Model}
\label{sec:model}

\label{sec:intro}
Denote the input acoustic feature sequence $\mathbf{x} = (x_1, x_2,...,x_T)$.  
In all of our models, the input sequence is passed through a multi-layer bidirectional long short-term memory (LSTM) network~\cite{hochreiter1997long}. 
Let the intermediate representation at the $i^{th}$ BiLSTM layer (that is, the output hidden state sequence after the $i^{th}$ layer) be $\mathbf{h}^i = (h_{1}^{i}, h_{2}^{i},...,h_{T}^{i})$.

\subsection{Connectionist Temporal Classification}
\label{subsec:ctc}
Connectionist temporal classification (CTC) is an approach for sequence labeling that uses a neural $N$-layer ``encoder'', which maps the input sequence $\mathbf{x}$ to a sequence of hidden states $\mathbf{h}^N$, followed by a softmax to produce posterior probabilities of frame-level labels (referred to as ``CTC labels") $p(\pi_t|h_t^N)$ for each label $\pi_t \in \mathcal{C}$.  
The posterior probability of a complete frame-level label sequence is given by the product of the individual frame posteriors:
\begin{equation}
p(\bm{\pi} = \pi_1, \pi_2, \ldots, \pi_T | \mathbf{x}) = \prod_t p(\pi_t|h_t^N)
\end{equation}
The CTC label set $\mathcal{C}$ consists of all of the possible true output labels plus a ``blank" symbol. Given a CTC label sequence, the hypothesized final label sequence is given by collapsing consecutive identical frame CTC labels followed by removing blank symbols. We use $B(\pi)$ to denote the collapsing function. 
All of the model parameters are learned jointly using the CTC loss function, which is the log posterior probability of the training label sequence $\textbf{z} = z_1, z_2 \ldots, z_L$ given input sequence $\mathbf{x}$,
\begin{eqnarray}
\log p(\mathbf{z} | \mathbf{x}) &=& \log \sum_{\bm{\pi} \in B^{-1}(\mathbf{z})} p(\bm{\pi} | \mathbf{x})\\
&=& \log \sum_{\bm{\pi} \in B^{-1}(\mathbf{z})} \prod_t p(\pi_t | h_t^N)
\end{eqnarray}
Model parameters are learned using gradient descent; the gradient is computed via the forward/backward technique~\cite{graves2006connectionist}.

\subsection{Baseline Model}
\label{subsec:baseline}

With a multi-layer bidirectional LSTM (BiLSTM) as our ``encoder'', we obtain a number of layer-specific hidden state sequences, $\mathbf{h}^1, \mathbf{h}^2 ,..., \mathbf{h}^N$, where $N$ is the number of layers in the BiLSTM.  The final encoder layer is followed by a softmax to produce label posteriors.  In this paper we use subword-level labels\footnote{By ``subword" here we refer specifically to sequence of {\it characters} contained within a word.}
which offer a good tradeoff between open-vocabulary decoding and lexical constraints.  These types of output labels have been used recently to build state-of-the-art open-vocabulary systems in speech recognition~\cite{chiu2017state} and machine translation~\cite{wu2016google}, specifically based on ``wordpieces'' constructed with byte pair encoding algorithm~\cite{sennrich2015neural,zenkel2017subword}.

\subsection{Hierarchical Multitask Training}

\begin{figure}[ht!]
\includegraphics[scale=0.65]{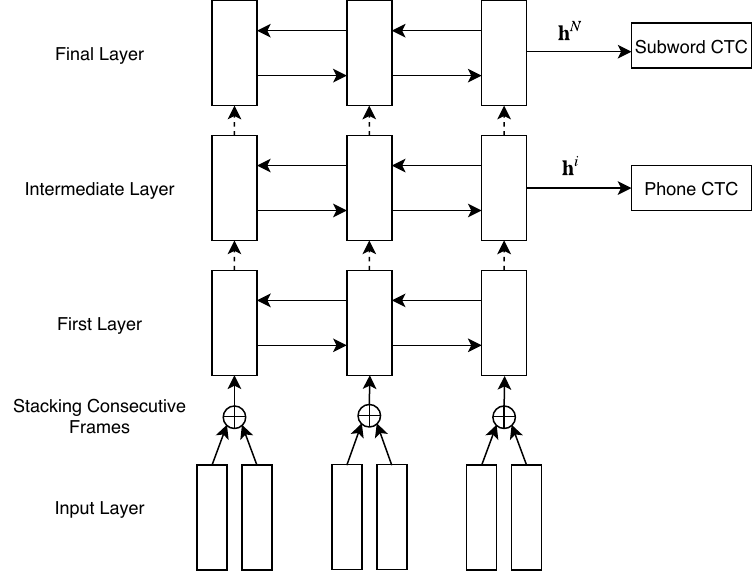}
\caption{A sketch of the hierarchical multitask learning model. Pairs of consecutive input frames are concatenated for a reduction in time resolution at the input. The Subword CTC loss is applied to the final layer. The Phone CTC loss is applied to an intermediate BiLSTM layer.}
\label{fig:model}
\end{figure}

Our primary objective is the subword-level CTC loss, applied to the softmax output after the final ($N^{th}$) encoder layer.
In our hierarchical multitask approach, shown in \figureref{model}, we add a second softmax after the $i^{th}$ encoder layer, to which we apply a phone-level CTC loss.  This softmax layer is added in parallel to the main model, and is used only at training time.  At test time only the main subword-level CTC model is used.  The choice of layer $i$ for the phone loss is a tunable hyperparameter.
The multitask loss is a convex combination of the two loss functions, using a tunable interpolation parameter $\lambda$: 
\begin{equation}
L = \lambda L_{\text{subword}}(\mathbf{h}^N, \mathbf{z}) + (1 - \lambda) L_\text{phone}(\mathbf{h}^i, \mathbf{z})
\end{equation}
While we use a phone-level auxiliary task here, the hierarchical multitask learning idea is general and can be used with any relevant intermediate loss, or multiple intermediate losses (as in~\cite{toshniwal2017mtl}).

\section{Experimental Setup}
\label{sec:experiment}

\subsection{Data}
We train all models on the Switchboard corpus (LDC97S62) \cite{godfrey1992switchboard}, which contains roughly 300 hours of conversational telephone speech. Following the Kaldi recipe~\cite{povey2011kaldi}, we reserve the first 4K utterances as a development set. Since the training set has many repetitions of short utterances (like ``uh-huh''), we remove duplicates beyond a count threshold of 300. The final training set has about 192K utterances. For evaluation, we use the HUB5 Eval2000 data set (LDC2002S09)~\cite{ld20022000}, consisting of two subsets: Switchboard (SWB), which is similar in style to the training set, and CallHome (CH), which contains conversations between friends and family.
For input features, we use 40-dimensional log-mel filterbank features along with their deltas, with per-speaker mean and variance normalization. 
For obtaining the phone labels, we map the words to their canonical pronunciations using the lexicon obtained from running the Kaldi Switchboard training recipe.

\subsection{Model Details and Inference}
Our encoder is a 5-layer bidirectional LSTM network~\cite{graves2005bidirectional}, with 320 hidden units in each direction. A dropout~\cite{srivastava2014dropout} mask has been applied on the output of each RNN layer~\cite{zaremba14rnn} (with dropout rate 0.1). 
We concatenate every two consecutive input vectors (as in \cite{sak2015fast, zweig2017advances}), which reduces the time resolution by a factor of two and speeds up computation.
For inference, we use greedy decoding with no language model.

We use a fixed vocabulary of 1000 wordpieces, which includes all of the characters to ensure open-vocabulary coverage. This vocabulary is generated using a variant of the byte pair encoding algorithm~\cite{sennrich2015neural} implemented in the SentencePiece library by Google.\footnote{\url{https://github.com/google/sentencepiece}}

\subsection{Training Details}
\label{sec:training}
We bucket training data by utterance length into 5 buckets, restricting utterances within a minibatch to come from a single bucket for training efficiency.
Different minibatch sizes are used for different buckets. For the shortest utterances a larger batch size of 128 is used while for the longest sequences the batch size is reduced to 32. 
We evaluate our models every 500 updates (3006 updates per epoch), measuring the word error rate (WER) on development data.
We use Adam \cite{kingma2014adam} for optimization with a fixed learning rate of 0.001 for the first 25K updates. 
Subsequently, the learning rate is reduced by a factor of 2 in case the current development WER is worse than the worst development WER of the previous 3 checkpoints. 
Training stops when there is no improvement in development WER for ten consecutive checkpoints. Early stopping on development set WER has been used to choose the final model for evaluation; that is, we use the model checkpoint with the best development set WER for final evaluation.
All models are trained on a single GPU using TensorFlow \texttt{r1.4} \cite{abadi2016tensorflow}.
\section{Results}
For our first experiment, we establish that our models are comparable to prior work and that hierarchical multitask training improves performance over a baseline CTC model. The baseline model corresponds to fixing $\lambda = 1$ (that is, using the subword CTC loss only). We compare this baseline to a model trained with a multitask objective with $\lambda = 0.5$ and the auxiliary phone CTC loss applied on the third layer ($i = 3$) of the 5-layer encoder. 
The results are given in \tableref{baseline}, along with several other recently published results in the same setting of lexicon-free recognition without a language model on the same test sets.  We observe a significant performance boost when including the auxiliary phone CTC loss. Our models are competitive with recently published results. Zenkel {\em et al.}~\cite{zenkel2017subword} report performance improvements using larger subword vocabularies, which we have not experimented with here. Zeyer {\em et al.}~\cite{zeyer2018improved} use an attention model with a novel layer-wise pretraining scheme.

We next analyze several aspects of our approach.
\begin{table}[t]
\small
\begin{center}
\caption{Word error rates (\%) of %baseline and proposed hierarchical multitask 
several models on the Switchboard development set and on Eval2000. SWB, CH = Switchboard, CallHome partitions of the Eval2000 corpus. Enc-Dec = Encoder-decoder.}%, CTC = Connectionist Temporal Classification.}

\label{tab:baseline}
\begin{tabular}{lcccc}
\toprule
Model & Dev  & \multicolumn{3}{c}{Eval2000} \\[0.1cm]
& & SWB & CH & full\\
\toprule
Our models & & & & \\
\hspace{0.1in}Baseline ($\lambda = 1.0$) & 25.5 & 21.5 & 33.8 & 27.7 \\
\hspace{0.1in}Multitask ($\lambda = 0.5$, $i = 3$) & 21.5 & 18.6 & 30.8 & 24.7  \\
\midrule
Enc-Dec & & & & \\
\hspace{0.1in}Zeyer {\em et al.}~\cite{zeyer2018improved} & - & \textbf{13.1} & 26.1 & 19.7 \\
\midrule
%Enc-Dec + Multitask & & & & \\
%\hspace{0.1in}Toshniwal {\em et al.}~\cite{toshniwal2017mtl} & 24.1 & 23.1 & 40.8 & 32.0 \\
%\midrule
CTC & & & & \\
%\hspace{0.1in}Maas {\em et al.}~\cite{maas2015lexicon} & - & 38.0 & 56.1 & 47.1 \\
\hspace{0.1in}Zweig {\em et al.}~\cite{zweig2017advances} & - & 24.7 & 37.1 & - \\
%\hspace{0.1in}Zenkel {\em et al.}~\cite{zenkel2017comparison} & - & 30.4 & 44.0 & 37.2 \\ 
\hspace{0.1in}Zenkel {\em et al.}~\cite{zenkel2017subword} & - & 17.8 & 30.4 & - \\
\hspace{0.1in}Sanabria {\em et al.}~\cite{sanabria2018hierarchical} & - & 17.4 & 28.5 & -\\
\hspace{0.1in}Audhkhasi {\em et al.}~\cite{audhkhasi2017building} & - & 14.6 & \textbf{23.6} & - \\
\midrule
CTC + Multitask & & & & \\
\hspace{0.1in}Sanabria {\em et al.}~\cite{sanabria2018hierarchical} & - & 14.0 & 25.5 & -\\
\bottomrule
\end{tabular}
\end{center}
\end{table}

\subsection{Effect of the Interpolation Weight}
\label{sec:interpolate}

\begin{figure}[ht!]
\begin{center}
\includegraphics[scale=0.55]{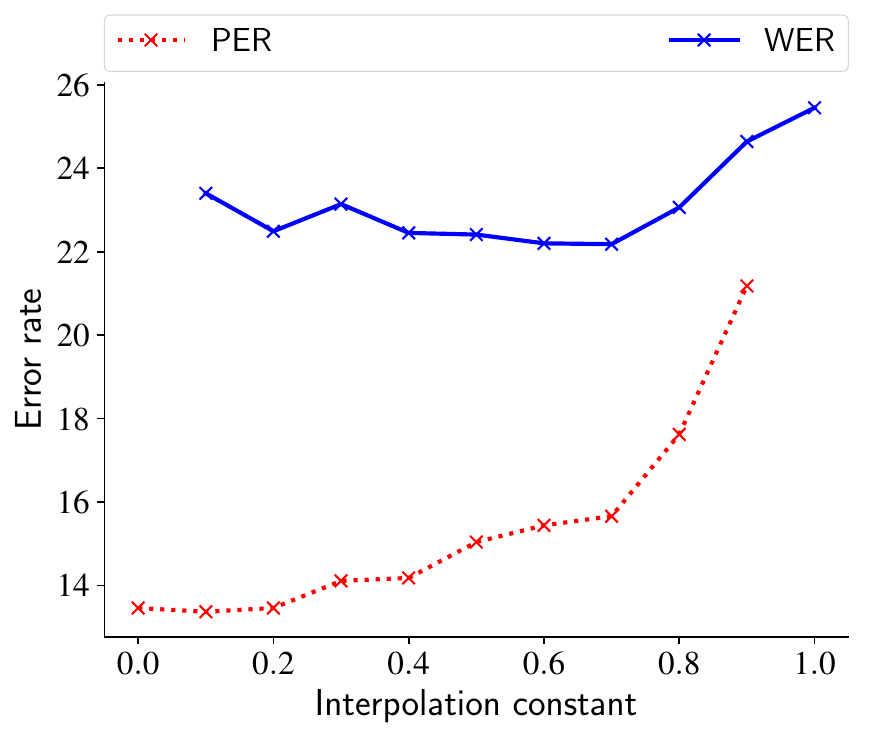}
\caption{Effect of varying the interpolation constant $\lambda$ on the development set word error rate (WER, \%) and phonetic error rate (PER, \%).}
\label{fig:interpolation}
\end{center}
\end{figure}

Next we explore how the mixing weight between the two terms in the multitask loss affects performance. We fix the auxiliary phone CTC loss on the fourth layer of the encoder ($i = 4$) and investigate the dependence of performance on the multitask interpolation constant $\lambda$.  
Development set results are shown in~\figureref{interpolation}.  
The multitask model with any value of $\lambda \neq 1$ outperforms the baseline ($\lambda = 1$), with the best performance attained at $\lambda = 0.7$. 

While phonetic recognition is not a goal of our work, it is interesting to study how the auxiliary model performs, and whether there is a correlation between performance of the auxiliary and main models.
We therefore also plot the phonetic error rate (PER) of the models as well.  
We observe a mostly monotonic improvement in PER with decreasing $\lambda$, that is with increasing weight on the phone CTC loss. 
There is a slightly worse performance for $\lambda = 0$ (13.5\%) compared to $\lambda = 0.1$ (13.4\%). 
This suggests that, if we view the phonetic recognition task as the primary task, additional supervision from the higher-level subword CTC task might help for a suitably small interpolation constant, although the effect is at best very small in our setting. Note that there is no single interpolation constant that optimizes performance for both tasks.

\subsection{Position of Auxiliary Loss}
\label{sec:position}
We next investigate the effect of the position of the auxiliary phone CTC loss function.  We fix $\lambda = 0.5$ and vary $i$, the layer to which the phone CTC loss is applied.  The development set results are given in the right-most column (the 100\% column) of \tableref{partial_wer}. All models match or outperform the baseline ($\lambda = 1.0$), with $i = 3$ giving the best WER. We obtain better performance for $i = 3,4$ than for $i = 5$, which is consistent with the previously observed benefit of applying the auxiliary loss to intermediate representations, rather than to the final output representation as in standard multitask learning~\cite{toshniwal2017mtl}.  We also note that $i = 1$ produces considerably worse performance than $i = 2, 3, 4, 5$.
We again also evaluate the phonetic recognition output, in terms of phonetic error rate (PER), produced by the auxiliary phone CTC model in our multitask models; see the 100\% column in \tableref{partial_per}. There is a consistent, fairly large improvement in PER as the phone CTC loss is placed on higher layers.  This is in contrast with the WER, which has a trough at intermediate values of $i$.  In other words, there is no single placement of the auxiliary loss that is best for both tasks.

These results -- in particular, the worse WER at $i = 1$ and the monotonic improvement in PER with $i$ -- may seem to be in contrast with earlier work analyzing learned neural representations~\cite{belinkov2017analyzing}, which reported that the first layer of a deep CTC recognizer (specifically, DeepSpeech 2~\cite{amodei2016deep}) is most useful for phonetic classification.  However, in this earlier work the model was trained only with the final character-level loss, and the phonetic classification tasks were trained on the resulting fixed representations.  We therefore do not consider these results to be conflicting.  This contrast does, however, indicate that it would be challenging to attempt to find the optimal layer for an auxiliary loss using an offline test on the baseline model.

\begin{table}[hb!]
\begin{center}
\caption{Development set word error rate (WER, \%) as the training set size is varied (10\% to 100\% of the standard 300-hour training set) and as the position of the auxiliary phone CTC loss is varied.  ``X'' indicates that the model did not converge.}
\label{tab:partial_wer}
\begin{tabular}{ lrrrr } 
\toprule
Model & 10\% & 25\% & 50\% & 100\% \\
\toprule
Baseline ($\lambda = 1.0$) & X & 39.7 & 29.1 & 25.5 \\
\midrule
Multitask ($\lambda = 0.5$) & & & & \\
~~~~$i = 1$~~~~~ & 48.4 & 36.1 & 28.6 & 25.5 \\
~~~~$i = 2$ & \textbf{44.7} & 35.3 & 26.8 & 23.0 \\
~~~~$i = 3$ & 45.6 & 33.4 & \textbf{25.7} &  \textbf{21.5} \\
~~~~$i = 4$ & \textbf{44.7} & 33.7 & 26.5 & 22.4 \\
~~~~$i = 5$ & \textbf{44.7} & \textbf{33.2} & 26.8 & 22.8 \\
\bottomrule
\end{tabular}
\end{center}
\end{table}

\begin{table}[hb!]
\begin{center}
\caption{Development set phonetic error rate (PER, \%) of the multitask model (with $\lambda = 0.5$) as the training set size is varied (10\% to 100\% of the standard 300-hour training set) and as the position of the phone CTC loss is varied.}
\label{tab:partial_per}
\begin{tabular}{ lrrrr } 
\toprule
Model & 10\% & 25\% & 50\% & 100\% \\
\toprule
$i = 1$~~~~~ & 47.5 & 41.1 & 37.1 & 36.2 \\
$i = 2$ & 35.0 & 31.0  & 25.2 & 22.8 \\
$i = 3$ & 31.5 & 23.6 & 19.9 & 17.0 \\
$i = 4$ & 27.9 & 21.8 & 16.8 & 15.0 \\
$i = 5$ & \textbf{26.8} & \textbf{19.7} & \textbf{15.6} & \textbf{13.2} \\
\bottomrule
\end{tabular}
\end{center}
\end{table}

\begin{center}
\begin{figure}[ht!]
\includegraphics[scale=0.55]{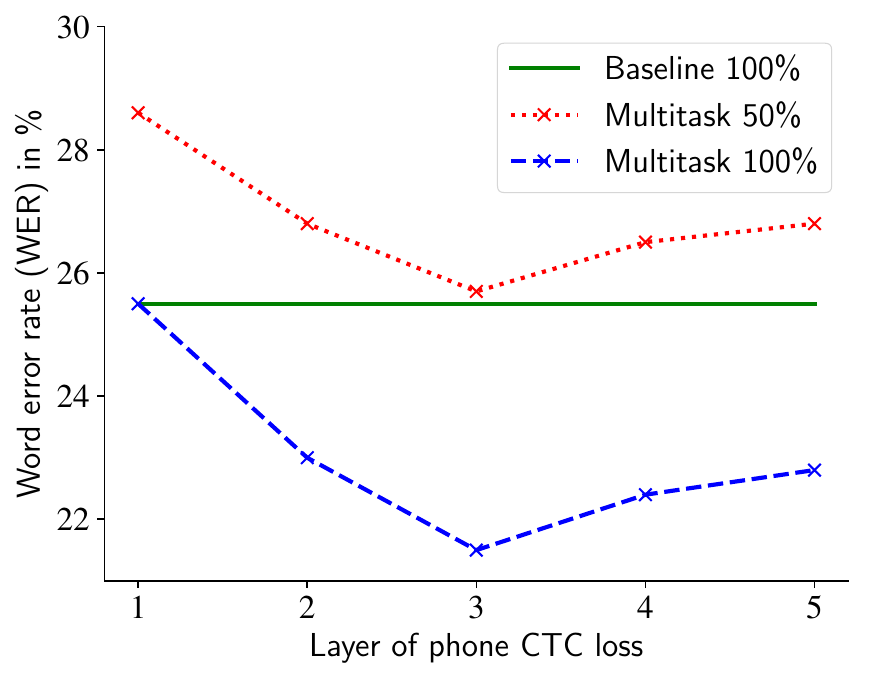}
\caption{Development set word error rate as the position of the phone CTC loss (value of $i$) is varied, when training on 50\% of the training set or the full (100\%) training set. For reference, we also plot the result of the baseline model trained on the full data set (``Baseline 100\%''). The complete results are given in \tableref{partial_wer}.}
\label{fig:partial_wer}
\end{figure} 
\end{center}

\subsection{Reduced Training Data}
\label{sec:partial}
We next investigate the effect of hierarchical multitask training in lower-data settings, specifically with only 10\%, 25\%, and 50\% of the full 300-hour Switchboard training set.  
To obtain a fraction of the full training set, we sample the same fraction of data from each of the five buckets which have utterances of different length distributions (\sectionref{training}). 
This is done to roughly maintain the same utterance length distribution in the partial training data.  We fix $\lambda = 0.5$ and vary $i = 1, 2, 3, 4, 5$.  We also train the baseline model ($\lambda = 1.0$) on all of the partial training sets.

The resulting word error rates (WER) are given in the 10\%, 25\% and 50\% columns of \tableref{partial_wer}.
For the smallest training set (10\%), the baseline model ($\lambda = 1.0$) fails to converge, while all of the multitask models do converge. This is in line with prior results~\cite{toshniwal2017mtl} indicating that hierarchical multitask learning not only boosts performance of end-to-end ASR systems, but also eases optimization. 
As in the full data setting (100\% column of \tableref{partial_wer}), all of the multitask models outperform the baseline ($\lambda = 1.0$), and the improvement is greater for $i = 2,3,4,5$ than for $i = 1$.
Interestingly, in the lowest-data settings (10\% and 25\%), unlike in the 50\% and 100\% data setting, there does not appear to be a benefit when placing the auxiliary task on intermediate BiLSTM layers: The model with $i = 5$ matches or slightly outperforms $i = 1, 2, 3, 4$ in the low 10\% and 25\% training data settings.  We add a graphical comparison between the 50\% data and 100\% data setting in \figureref{partial_wer}. We note that at the 50\% data setting, the best hierarchical multitask model is able to almost match the baseline model using the {\it full} data set (25.7\% vs. 25.5\% WER).  This is consistent with the intuition that multitask training should improve generalization.

As before, we present results on the auxiliary phonetic recognition task in the partial data setting in \tableref{partial_per}. A regular trend is observed, with PER always improving significantly as the phone CTC loss is placed on higher layers (moving down in \tableref{partial_per}). For the 10\% and 25\% settings, $i = 5$ is the auxiliary loss position that maximizes performance on both tasks.

\subsection{Pretraining vs.~Multitask Training}
\label{sec:pretrain}
 Hierarchical multitask learning is intended to improve the learnt intermediate representations via additional auxiliary supervision at intermediate layers.  Another way of potentially achieving this effect is by pretraining the lower model layers with the auxiliary loss such as the phonetic loss. For example, Audhkhasi {\em et al.}~\cite{audhkhasi2017direct, audhkhasi2017building} reported significant improvements by pretraining a whole-word BiLSTM CTC network with a phone CTC loss, as well as more modest improvements in a hierarchical setting where all but the last BiLSTM layer were pretrained and the full model was then trained with multitask (word + phone CTC) learning.

This approach bears resemblance to transfer learning for neural models, which is often used in computer vision~\cite{oquab2014learning}, and also sometimes in natural language processing (NLP)~\cite{PetersELMo2018} and speech recognition~\cite{audhkhasi2017direct}. In a typical transfer learning approach, a model is initially pretrained on a basic task (e.g., language modelling in NLP or image classification in computer vision). The learned weights of the first few layers are then used to initialize the lower layers of models for other tasks (such as sentiment classification or question answering for NLP).  This approach is a form of pretraining, but with the pretraining typically done on a different data set from that of the final task.

In our pretraining experiments, we start out by training a 3-, 4- or 5-layer model with only the phone CTC loss ($\lambda = 0.0$ for $i = 3, 4, 5$).  For this pretraining phase, we follow an optimization scheme identical to that of \sectionref{experiment}, with early stopping based on development set phonetic error rates instead of word error rates. After this pretraining, we initialize a 5-layer subword CTC model with the learnt phone CTC weights for the first $i$ layers, and randomly for the remaining layers\footnote{The final softmax layer in the subword-level CTC model is randomly initialized regardless of $i$.}, and train all the layers with only the subword-level CTC loss ($\lambda = 1.0$) until convergence (we call these models ``Pretrain'').

Finally, we also consider combining pretraining and multitask learning.  In this approach we begin as in the ``Pretrain'' models, by pretraining $i$ layers with a phone CTC loss and then initializing a 5-layer model with the pretrained weights on the lower layers and random weights on the remaining layers.  However, in the combined approach (``Pretrain + Multitask''), we keep the phone softmax on layer $i$ and train with a multitask phone CTC + subword CTC loss with $\lambda = 0.5$.

We present our results in \tableref{pretrain}. Both ``Multitask'' and ``Pretrain'' improve over the baseline for all values of $i$, with the best development WER at $i = 3$ and $i = 4$ respectively. While the multitask models outperform the pretrained ones on the development set, they have nearly identical performance on the test sets. However, for all values of $i$, the ``Pretrain + Multitask'' model outperforms the corresponding ``Multitask'' and ``Pretrain'' models. For all three model types, a hierarchical CTC model ($i = 3, 4$) outperforms the vanilla multitask setting ($i = 5$).  The final best-performing model is ``Pretrain + Multitask'' with $i=4$, obtaining a 3.6\% absolute reduction in word error rate on the Switchboard test set and a 3.1\% absolute reduction on CallHome.\footnote{It may be possible to further improve results by also tuning $\lambda$.}

As in our previous experiments, we also examine the development set phonetic error rates in \tableref{pretrain}. For ``Pretrain'', the PER is calculated after pretraining on the Phone CTC loss, and before the actual training the subword-level CTC model. For ``Multitask'' and ``Pretrain + Multitask'', the final trained phone CTC model is evaluated. We observe a consistent trend:  the best PERs are obtained with $i = 5$ across all model types, and for a given $i$ the ``Pretrain + Multitask'' model performs best, followed by ``Pretrain'' and then ``Multitask''.

\begin{table}[hb!]
\small
\begin{center}
\caption{A comparison of pretraining and multitask learning on our Switchboard development set and evaluation of the tuned models on Eval2000.  SWB, CH = Switchboard, CallHome partitions of the Eval2000 corpus. All of the ``Multitask'' models are trained with $\lambda = 0.5$. $i$ denotes the position of the phone CTC loss throughout the experiment. PER, WER refer to development set phonetic error rate and word error rate. PER for the ``Pretrain'' models has been calculated right after pretraining (and before training the subword CTC model).}
\label{tab:pretrain}
\begin{tabular}{lccccc}
\toprule
Model & \multicolumn{2}{c}{Dev} & \multicolumn{3}{c}{Eval2000} \\[0.1cm]
& PER & WER & SWB & CH & full\\
\toprule
Baseline ($\lambda = 1.0$) & - & 25.5 & 21.5 & 33.8 & 27.7 \\
\midrule
Multitask & & & & \\
\hspace{0.1in}$i = 3$ & 17.0 & \textbf{21.5} & 18.6 & 30.8 & 24.7  \\
\hspace{0.1in}$i = 4$ & 15.0 & 22.4 & - & - & - \\
\hspace{0.1in}$i = 5$ & 13.2 & 22.8 & - & - & - \\
\midrule
Pretrain & & & & \\
\hspace{0.1in} $i = 3$ & 15.2 & 22.2 & - & - & - \\
\hspace{0.1in} $i = 4$ & 13.5 & \textbf{22.0} & 18.6 & 30.7 & 24.7  \\
\hspace{0.1in} $i = 5$ & 13.1 & 23.7 & - & - & -  \\
\midrule
Pretrain + Multitask & & & & \\
\hspace{0.1in} $i = 3$ & 14.9 & 21.4 & - & - & - \\
\hspace{0.1in} $i = 4$ & 12.9 & \textbf{21.2} & 17.9 & 30.7 & 24.3 \\
\hspace{0.1in} $i = 5$ & 12.8 & 23.3 & - & - & - \\
\bottomrule
\end{tabular}
\end{center}
\end{table}

\begin{figure*}[t!]
\small
\begin{center}
\begin{tabular}{ |l|c|c|c|c|c|c|c|c|c|c|c|c|c|c|c|c|c|c|c|c|c|c| } 
\hline
Baseline subword 
& \_ & \_ & \_ & for & \_ & \_ & \_ & the & the & \_ & \_ & \_ & \_ & \_ & last & \_ & \_ & \_ & \_ & \_ & \_ & \_ \\
\hline
Multitask subword 
& \_ & \_ & for & \_ & \_ & \_ & \_ & the & \_ & \_ & \_ & \_ & \_ & last & \_ & \_ & \_ & \_ & \_ & \_& \_ & \_ \\
\hline
Multitask phone 
& f & \_ & \_ & er & \_ & dh & dh & dh & ah & \_ & \_ & l & l & \_ & \_ & ae & \_ & \_ & \_ & s & \_ & t \\
\hline
Ground Truth & \multicolumn{5}{|c|}{for} & \multicolumn{3}{|c|}{the} & \multicolumn{14}{|c|}{last} \\
\hline
\multicolumn{23}{c}{}\\
\multicolumn{23}{c}{}\\
\hline
Baseline subword  
& \_ & the & \_ & \_ & \_ & \_ & \_ & \_ & \_ & se & \_ & \_ & co & co & co & \_ & \_ & nd & nd & \_ & \_ & \_ \\
\hline
Multitask subword 
& the & \_ & \_ & \_ & \_ & \_ & \_ & \_ & se & \_ & \_ & \_ & \_ & \_ & \_ & \_ & co & nd & nd  & \_ & \_ & \_ \\
\hline
Multitask phone 
& dh & ah & \_ & s & s & \_ & \_ & \_ & \_ & \_ & eh & \_ & \_ & \_ & k & \_ & \_ & ax & n & n & d &  \_\\
\hline
Ground Truth & \multicolumn{2}{|c|}{the} & \multicolumn{20}{|c|}{second} \\
\hline
\end{tabular}
\end{center}
\caption{Alignment of per-frame CTC outputs for the baseline and proposed model, compared with ground-truth alignments. The ``\_'' token refers to the CTC blank symbol.}
\label{fig:alignment}
\end{figure*}

\subsection{Example Alignments}
\label{sec:alignment}
Finally, we examine the frame-level CTC outputs for the baseline model ($\lambda = 1.0$) and a multitask model ($\lambda = 0.5, i = 4$) and compare them in \figureref{alignment}. These sequences are obtained from a greedy decoding of the subword-level output layer of the two models, and the phone output layer of the multitask model. We compare our outputs with the Mississippi State ground-truth word alignments~\cite{deshmukh1998resegmentation}.

We notice that the multitask model generally (but not always) outputs its output token one frame (20ms) earlier than the baseline model does, perhaps indicating its higher confidence in prediction.  Alignments for 128 development set examples can be found at \url{http://martiansideofthemoon.github.io/ctcalign/}.

\section{Conclusion}
We take a step forward in end-to-end neural speech recognition by presenting a detailed analysis of a hierarchical multitask learning framework for CTC-based recognition. This is a general approach for deep end-to-end neural models, where we use some prior knowledge about intermediate tasks to design auxiliary tasks applied at an intermediate layer during training.  In our case we use phonetic recognition as an intermediate task along the way to word (or in this case, wordpiece) recognition.

The key takeaways from our paper are the following: 1) Multitask learning consistently improves performance. In our higher-resource settings, hierarchical multitask learning is better than the vanilla multitask approach, whereas the vanilla multitask approach performs better in the lower-resource experiments 2) Pretraining with the auxiliary task and multitask learning both improve performance in a hierarchical setting, but the best results are obtained by combining the two approaches, again in a hierarchical fashion 3) In general there is no single task interpolation constant that gives the best results on both the main task (word recognition) and auxiliary (phonetic recognition) task.  The approach therefore is helpful for the main task but at best only very slightly for the lower-level task.  It is worth further considering in future work, however, whether higher-level tasks can also help lower-level ones. 4) Multitask learning assists optimization as well as generalization, reconfirming prior work.  

Our best-performing models, which combine pretraining and multitask learning, achieve 3.4\% absolute improvement in word error rate on Eval2000 over the baseline subword-level CTC model.  This is a larger improvement, over a stronger baseline, than has been found previously for encoder-decoder speech recognition models on the same data~\cite{toshniwal2017mtl}, and using a single auxiliary task unlike the combination of multiple tasks in~\cite{toshniwal2017mtl}.  Additional auxiliary tasks may further improve CTC-based models as well.

Future work includes developing ways of pre-determining the best design and hyperparameter choices (auxiliary task, interpolation constant, and position of auxiliary loss) to avoid training and testing many complete multitask models; further study of hierarchical multitask training on an even larger range of data set sizes; and exploration of additional auxiliary tasks. 

\bibliographystyle{IEEEbib}
\bibliography{strings,refs}

\end{document}